\documentclass[letterpaper, 10 pt, conference]{ieeeconf}  

\IEEEoverridecommandlockouts                              

\usepackage{booktabs}
\usepackage{graphicx}                                                         
\usepackage{float} 
\usepackage{multirow}
\usepackage{multicol}
\usepackage{amsmath}  

\usepackage{stfloats}  

\usepackage {hyperref}  

\title{\LARGE \bf
FCRF: Flexible Constructivism Reflection for Long-Horizon Robotic Task Planning with Large Language Models
}

\author{Yufan Song$^{1\dag}$, Jiatao Zhang$^{1\dag}$, Zeng Gu$^{2}$, Qingmiao Liang$^{2}$, Tuocheng Hu$^{1}$, Wei Song$^{1\star}$, and Shiqiang Zhu$^{1\star}$
\thanks{$^{1}$Zhejiang University, Hangzhou, China}%
\thanks{$^{2}$Hangzhou Institute for Advanced Study, University of Chinese Academy of Sciences, Hangzhou, China}
\thanks{$^{\star}$Corresponding emails: \texttt {\{weisong-rob,sqzhu\}@zju.edu.cn}}%
\thanks{$^{\dag}$ Contribute equally to this work.}
}

\begin{document}

\maketitle
\thispagestyle{empty}
\pagestyle{empty}

\begin{abstract}


Autonomous error correction is critical for domestic robots to achieve reliable execution of complex long-horizon tasks. Prior work has explored self-reflection in Large Language Models (LLMs) for task planning error correction; however, existing methods are constrained by inflexible self-reflection mechanisms that limit their effectiveness. Motivated by these limitations and inspired by human cognitive adaptation, we propose the Flexible Constructivism Reflection Framework (FCRF), a novel Mentor-Actor architecture that enables LLMs to perform flexible self-reflection based on task difficulty, while constructively integrating historical valuable experience with failure lessons. We evaluated FCRF on diverse domestic tasks through simulation in AlfWorld and physical deployment in the real-world environment. Experimental results demonstrate that FCRF significantly improves overall performance and self-reflection flexibility in complex long-horizon robotic tasks. 

\end{abstract}

\section{INTRODUCTION}

Robotic task planning constitutes a fundamental capability for high-level decision-making, enabling robots to generate executable action sequences through environmental perception, capabilities, and task objectives~\cite{guo2023recent}. Domestic robots increasingly assume critical roles in assistive human living, which interact with various objects and generate longer action sequences, often leading to numerous errors that accumulate over time~\cite{zachiotis2018survey}. This characteristic highlights the need for autonomous error correction to ensure stable execution of complex long-horizon task planning.


In recent years, powered by massive data training, rapidly developed LLMs possess encyclopedic world knowledge and situated language comprehension capabilities, enabling their application in robotic task planning~\cite{ahn2022can,kannan2024smart,zhou2024isr}. 
LLMs enhance robotic planning and reasoning capabilities, allowing direct interaction with task instructions in natural language and improving overall task planning performance. This leads us to ask: can an LLM agent with self-reflection capabilities be designed to autonomously correct errors and optimize actions for domestic robots task planning, thereby better addressing the complex needs of long-horizon tasks?

Several existing studies have explored self-reflection mechanisms for error correction in the LLM task planning process. Foundational approaches include RETROFORMER~\cite{yao2023retroformer}, and Reflexion~\cite{shinn2024reflexion} based on ReAct~\cite{yao2022react} actor, processing reflection gradients through textual representations. 
Another series of works~\cite{yao2024tree,zhang2024agent,zhou2023language} combine tree search with reflection to deliberately seek a better solution. Recently, works like Expel~\cite{zhao2024expel} and AutoManual~\cite{chen2024automanual} build rule systems according to the task environment, leading to comprehensive reflection on previous errors of LLM agents.


\begin{figure}[t!]
    \centering
    \includegraphics[width=0.50\textwidth]{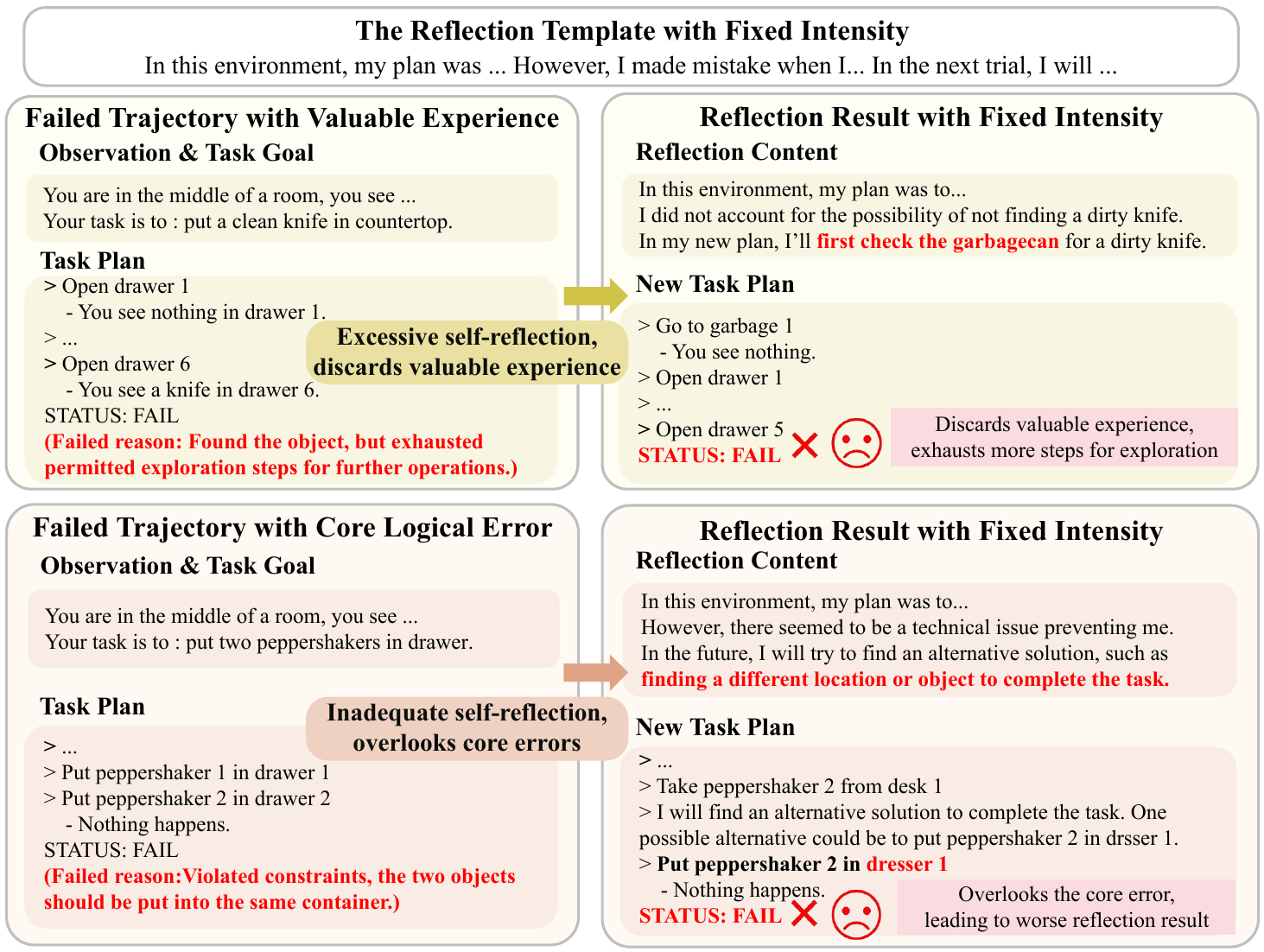}
    \caption{Illustration of the flexibility problem of self-reflection during long-horizon LLMs task planning. If LLMs always reflect at a fixed intensity, they may discard valuable experience in minor error scence or overlook core errors, leading to suboptimal reflection even worse planning results.}
    \label{fig1_motivation}
    \vspace{-3mm}
\end{figure}

Although previous methods show promising results, a key challenge in applying LLMs for self-reflection to complex long-horizon task planning is their lack of flexibility. Specifically, current frameworks typically employ a fixed reflection paradigm, failing to adjust the intensity of reflection according to the nature of the errors. In long-horizon robotic tasks, the severity of errors varies significantly across different trajectories. For example, some failures result from insufficient exploration, where most actions are correct and only minor adjustments are needed. In such cases, a light reflection suffices. In contrast, failures caused by fundamental logical errors require more intensive reflection and substantial modifications to the trajectory. Therefore, it is crucial for LLMs to integrate deeply with the current task trajectory and adaptively adjust the reflection process. As illustrated in Figure~\ref{fig1_motivation}, if LLMs reflect at a fixed intensity in complex long-horizon tasks, they risk discarding valuable experience or overlooking the root cause of errors, leading to suboptimal reflection and, ultimately, worse planning results.


Motivated by the challenges mentioned, we investigate the flexibility of LLMs' self-reflection in complex long-horizon robotic task planning. Humans similarly encounter varying degrees of error severity in long-horizon tasks. When errors occur, humans review their actions, analyze useful experience, and integrate external knowledge to refine both successes and failures, reflecting with appropriate intensity to correct the errors.
This characteristic extends to widely recognized constructivist learning theory~\cite{hein1991constructivist}, as recognized in educational psychology, underscores that effective learning integrates new and prior experience within the task context. For example, when human students make mistakes, a good educator first acknowledges the correct aspects of their work, then addresses specific errors by drawing on relevant knowledge from a broader system to guide the students. This helps students integrate new and old experience, fostering an appropriately intense reflection.


Inspired by the aforementioned process, we propose the Flexible Constructivism Reflection Framework (FCRF), an online approach enabling LLMs to perform flexible self-reflection based on task difficulty.
Our method draws from constructivist learning, using a Mentor-Actor LLM architecture for online reflection. The Actor LLM follows the ReAct~\cite{yao2022react} framework for action planning, while the Mentor LLM guides the reflection process for error correction. The Mentor summarizes successful experience from the trajectory of Actor and extracts failure lessons from corrected trajectories, maintaining a universal Lesson Pool. 
Focusing on the issue of flexibility, our framework includes a complexity assessment module, allowing the LLM to select reflection intensity tailored to evaluated task difficulty. 
Once reflection intensity is determined, valuable experience and failure lessons are constructively integrated into a new plan, guiding the Actor LLM in the planning of next trail.

To evaluate the performance of our framework, we perform experiments in AlfWorld~\cite{shridhar2020alfworld} household planning tasks. We divide all the tasks in the dataset into six categories according to their operation type.
The results show that our framework achieves the best performance on all tasks, and improves 31.2\% reflection flexibility, 25.0\% valuable experience recall and 63.3\% error correction precision. 
In summary, the contributions of our work are as follows:
\begin{enumerate}
\item {To the best of our knowledge, we are the first to define and research the flexibility problem of LLMs reflection for complex long-horizon robotic task planning.} 
\item {We propose a novel Flexible Constructivism Reflection Framework (FCRF), which dynamically determines the intensity of reflection based on task complexity and integrates feedback from task success or failure for adaptive and efficient self-reflection processes.}
\item {We introduce a Mentor-Actor reflection architecture, where the Mentor leverages a lesson pool, a generalized knowledge base across scenarios. This base may contain new knowledge for human, and supports multiple maintenance methods, including human knowledge injection and LLM-based summarization.} 
\item {We validate our proposed method in both AlfWorld simulation environments and real-world robotic experiments, demonstrating that our framework significantly improves task performance and adaptability under complex and diverse conditions.} 
\end{enumerate}

\section{RELATED WORKS}

\subsection{LLMs for Task Planning}

Several prior works have employed LLMs for task planning, on account of their inherent powerful reasoning and planning capabilities. 
For instance, classic LID~\cite{li2022pre} uses pre-trained GPT-2~\cite{radford2019language} as a general framework for interactive decision making, by converting policy inputs including observations, goals, and history into sequential data. These embeddings are then passed to a pre-trained policy network to predict actions. 
ReAct~\cite{yao2022react} combines reasoning and acting with language models for solving diverse language reasoning and decision making tasks. 
Several later works like CodeAsPolicy~\cite{liang2023code}, ProgPrompt~\cite{singh2023progprompt} and AdaPlanner~\cite{sun2024adaplanner}, consider the powerful programming capability of LLMs, propose to use programmatic code as the plan of LLMs.
Focusing on the combination of robots and LLMs, typical embodied robotic work SayCan~\cite{ahn2022can} extracts and leverages the knowledge within LLMs in physically-grounded tasks, constraining the model to propose natural language actions that are both feasible and contextually appropriate. 
A series of other works including LLMPlanner~\cite{song2023llm,wake2023chatgpt} also develop LLMs for use in robotic planning tasks.

\subsection{LLM Agents for Self-Reflection}

Developed from simple planning strategies, a series of later studies adopt self-reflection methods, using feedback to perform multistep planning and error correction of LLM agents. 
For example, improved on the ReAct reasoning framework, Reflexion~\cite{shinn2024reflexion} allows LLMs to reflect on their previous failures according to environmental feedback, forming an improved plan for the next attempt. 
Based on the Reflexion framework, recent work Expel~\cite{zhao2024expel} builds an offline learning process, the LLM agent gathers experience from a collection of training tasks through trial and error.
AutoManual~\cite{chen2024automanual} builds a well-organized understanding of the environment that can guide multitask planning effectively. 
Another series of works~\cite{yao2024tree,zhang2024agent,zhou2023language} combine tree search with reflection to seek a better solution to the task.

\section{PRELIMINARIES}

\subsection{Planning Framework}

A task can be defined as a tuple $\langle G, S, O, T, A \rangle$~\cite{lauri2022partially}, where $G$ represents the task goal, $S$ represents the set of all possible states, $O$ is the set of observations of task environment, $A$ is the set of possible actions, and $T$ is the transition function, formally formalized as $\mathrm{T} : S \times A \rightarrow S$, represents the environmental state changes because of actions. 
The objective is to find a plan $\pi$, in the form of a sequence of actions, that transitions from the initial state to the target state. 
There is currently no unified and strict definition of complex long-horizon tasks. Based on the summary of existing methods, we can generally define tasks with more than 10 steps or even longer, simultaneously interacting with a greater variety of items and environments, as complex long-horizon tasks~\cite{zhang2024fltrnn}.

\subsection{Self-Reflection Process in LLMs Planning}

Discussing LLMs task planning process with self-reflection and memory module $mem$, the self-reflection content for current unsuccessful planning trail $sr^{t} \in SR$ is generated by the LLM for self-reflection $M_{sr}$, while the reflection process can be described as $M_{sr}(sr^{t}|s^{t},mem^{t})$, meaning that the self-reflection content $sr^{t}$ of the current trail $t$ is generated by $M_{sr}$ based on the current state $s^{t}$, and persistent memory $mem^{t}$ lasts until the current trail including existing task trajectory. 

The self-reflection generated by $M_{sr}$ would be stored in the agent memory $mem$ and passed to planner LLM in the next trail, to generate better action sequence $\{a^{t+1}_{1},\dots,a^{t+1}_{i}\}$ as the new output of the planning task. The sequence of actions is generated by the strategy function $\Phi(a^{t+1}_{1},\dots,a^{t+1}_{i}|G,s^{t+1},o^{t+1},sr^{t})$, meaning that new action sequence $\{a^{t+1}_{1},\dots,a^{t+1}_{i}\}$ from $step_{1}$ to $step_{i}$  is generated based on the task goal $G$, the current agent state $s^{t+1}$, the current observation of task environment $o^{t+1}$, and the generated self-reflection content $sr^{t}$ of last failed trail. 

\begin{figure*}[!htp]
    \centering
    \includegraphics[width=0.7\textwidth]{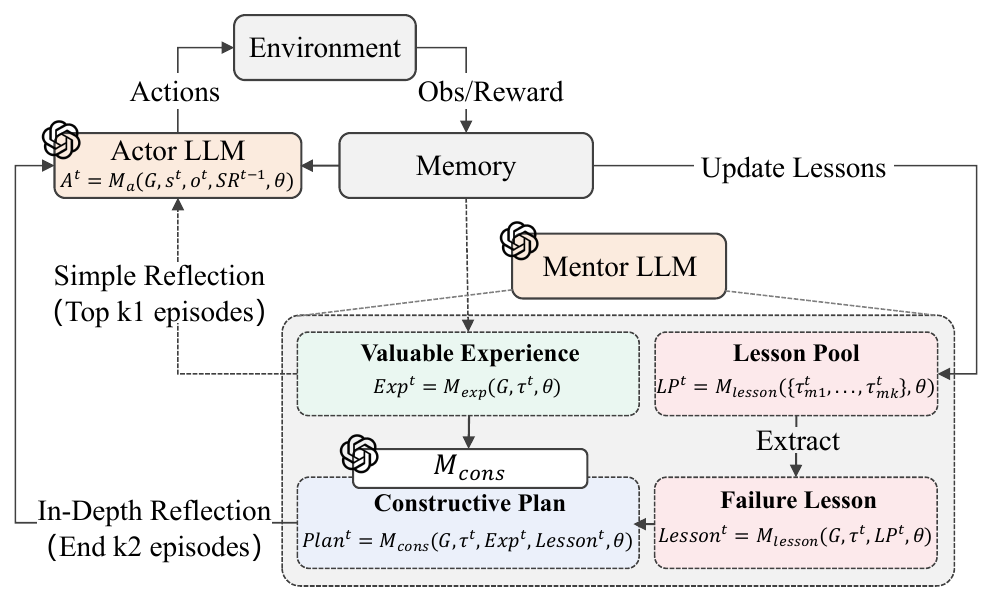}
    \caption{The framework of FCRF. The planning process is executed by the Actor LLM, the reflection process is executed by the Mentor LLM. During the self-reflection process, the difficulty level of the task is first assessed, according to which the Mentor flexibly selects reflection intensity, determine the proportion of simple experience retain and in-depth failure lessons extraction among all the reflection episodes. Combining the valuable experience and failure lesson, the Mentor performs a constructivism self-reflection to guide the next round of planning of the Actor. The reflection results and planning trajectories will be stored in the memory module for long-term management.}
    \label{framework}
    \vspace{-3mm}
\end{figure*}

\section{METHODOLOGY}

\begin{figure*}[!htp]
    \centering
    \includegraphics[width=0.99\textwidth]{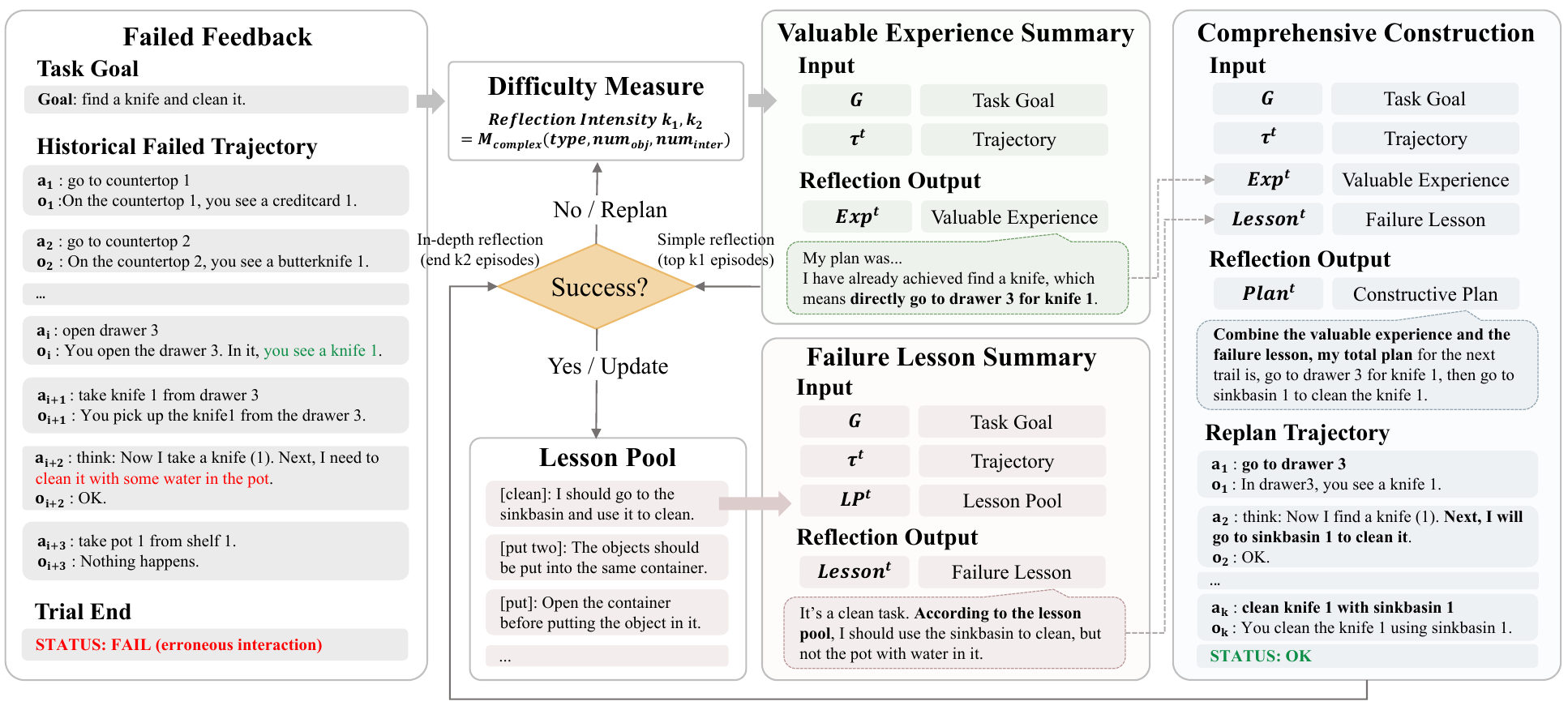}
    \caption{The detailed illustration of FCRF. The difficulty level of the task is first assessed, determining the number of episodes for simple and in-depth reflection. After the reflection intensity is determined, the detailed self-reflection process consists of the valuable experience summary process, the failure lesson summary process and the comprehensive construction process. Final output of the self-reflection process is a improved new plan for the next attempt.}
    \label{FCRF_detail}
    \vspace{-3mm}
\end{figure*}

In this section, we present the details of our framework Flexible Constructivism Reflection Framework (FCRF). As illustrated in Figure~\ref{framework}, our FCRF is a Mentor-Actor architecture that consists of four parts: an LLM as planning Actor, an LLM as self-reflection Mentor, the memory management module and the overall flexible reflection process. These modules will be introduced specifically in the following.

\subsection{LLM as Actor}

Following the architecture of the classic Reflexion method, our planning process is completed by an LLM prompted as an $Actor$, represented as $M_{a}$ in the following. $M_{a}$ takes task goal, current environmental observations and reflection text of previous failed task trails as inputs, and is specifically prompted to generate action sequence in text form as output. This planning process of $M_{a}$ can be described as:

\begin{equation}
A^{t}=M_{a}(G,s^{t},o^{t},SR^{t-1},\theta),
\label{Actor_LLM} 
\end{equation}
where $A^{t}=\{a^{t}_{1},\dots,a^{t}_{i}\}$ denotes the current action sequence, $G$ represents the task goal, $o^{t}$ represents the current environmental observations during trail $t$, $SR^{t-1}=\{sr^{t-k},\dots,sr^{t-1}\}$ represents the set of self-reflection among the past $k$ trials to consider about, and $\theta$ represents the parameters of $M_{a}$. 
The action sequence generated by Actor LLM $M_{a}$ would be stored in the memory module $mem$ for the following self-reflection process of the task, until the current task is successfully completed. 

\subsection{LLM as Constructivism Self-Reflection Mentor}

The constructivism self-reflection process of our framework is guided by an LLM prompted as a $Mentor$, represented as $M_{m}$, which contains three sub-modules $M_{exp}$, $M_{lesson}$ and $M_{cons}$. The reflection process is divided into three parts: the summary of valuable experience, the summary of failure lessons, and the comprehensive construction of the two above parts. The final generated reflection content will be added to the memory module $mem$ and can be called by the Actor LLM $M_{a}$ in subsequent trails for planning.

\textbf{The valuable experience summary process} is completed by the submodule $M_{exp}$. $M_{exp}$ analyzes the current trail trajectory planned by $M_{a}$, summarizing effective interaction attempts contained in the unsuccessful trail, then retains them as valuable experience and integrates them into the memory, for usage in the construction of subsequent reflection content.
The process can be described as:

\begin{equation}
Exp^{t}=M_{exp}(G,\tau^{t},\theta),
\label{experience} 
\end{equation}
where $G$ represents the task goal, $\tau^{t}$ represents the trajectory of current trail $t$, $\theta$ represents the parameters of $M_{exp}$, mainly embodies through prompt. 

\textbf{The failure lesson summary process} is completed by the submodule $M_{lesson}$, specifically divided into the preliminary maintenance of the mentor lesson pool and the failure lesson extraction process. 
The mentor lesson pool is expressed as $LP$, which can be regarded as a universal knowledge base of the task environment. 
During the lesson pool maintenance process, $M_{lesson}$ accesses the existing cross-task trajectories stored in memory online, summarizing scenario universal failure lessons from all trajectories that have been successfully corrected by the episode in which the current trail locates, then incrementally add the new lesson to $LP$. 
The failure lesson summary process can be described as:
\begin{equation}
LP^{t}=M_{lesson}(\{\tau^{t}_{m1},\dots,\tau^{t}_{mk}\},\theta),
\label{lesson_summarize} 
\end{equation}
where $LP^{t}$ represents the up-to-date lesson pool set up to trail $t$, $\{\tau^{t}_{m1},\dots,\tau^{t}_{mk}\}$ represents all $k$ successfully modified task trajectories in the current trail $t$, $\theta$ represents the parameters of $M_{lesson}$, mainly embodies through prompt. 

During the failure lesson extraction process, $M_{lesson}$ combines the current task goal to analyze the current trail task trajectory, then extracts a lesson tip that best fits the cause of the current task failure from the lesson pool maintained so far, as external information for failure reflection. The failure lesson will be used in the construction of subsequent reflection content together with valuable experience $Exp^{t}$.
The process of failure lesson extraction can be expressed as:
\begin{equation}
Lesson^{t}=M_{lesson}(G,\tau^{t},LP^{t},\theta),
\label{lesson_extract} 
\end{equation}
where $G$ represents the task goal, $\tau^{t}$ represents the trajectory of current trail $t$, $LP^{t}$ represents the up-to-date lesson pool set up to trail $t$, $\theta$ represents the parameters of $M_{lesson}$, mainly embodies through prompt.

After completing the valuable experience summary process and the failure lesson summary process, the constructor submodule $M_{cons}$ will execute \textbf{the comprehensive construction process} to comprehensively construct the successful and failed experience, integrate the final reflection result in the form of an improved plan. 
During the construction process, $M_{cons}$ integrates the summarized valuable experience $Exp^{t}$ with the newly injected $Lesson^{t}$, unifying them into a sequence of actions, which forms the new constructive plan to be executed in the next trail. 
The reflected new constructive plan will be stored in the $mem$ module and guide the planning process of $M_{actor}$ in the next trail. 
The comprehensive construction process can be described as:
\begin{equation}
Plan^{t}=M_{cons}(G,\tau^{t},Exp^{t},Lesson^{t},\theta),
\label{constructivism} 
\end{equation}
where $G$ represents the task goal, $\tau^{t}$ represents the trajectory of current trail $t$, $Exp^{t}$ and $Lesson^{t}$ respectively represents the valuable experience and the precisely required failure lesson summarized through trail $t$, $\theta$ represents the parameters of $M_{cons}$, mainly embodies through prompt.

\subsection{Memory Management}
During the task planning process, it is important for LLMs to maintain context memory. However, if all memory contents are included in prompts, it will lead to redundant prompts, increasing the memory burden of LLMs, even affect their performance. 
Motivated by the long-short-term memory mechanism of previous work Reflexion, we design the memory module $mem$ for comprehensive memory management, which is equivalent to an external buffer of LLMs. The $mem$ is divided into trajectory management module $Traj^{t}$ and reflection management module $Refl^{t}$, comprehensively manages the memory of task trajectories and reflection contents. 
The $Traj^{t}$ module is a short-term memory module storing the task trajectory $\tau^{t}$ of the current trail $t$. 
The $Refl^{t}$ module is a long-term memory module storing all failure self-reflection contents up to trail $t$, which can be represented as $\{sr^{1},\dots,sr^{t}\}$. The reflection contents are stored for targeted reading and calling by the Mentor LLM $M_{m}$ in the following trails.
Our memory mechanism $mem$ relieves the memory pressure of LLMs, improving the efficiency of memory management and reflection process.

\subsection{The Flexible Reflection Process}
As an important innovation, our framework adopts a flexible self-reflection process. For tasks with different difficulties, our method adopts reflection frameworks of different intensities. We design the model $M_{complex}$ to calculate the difficulty level of the task and thus determine the proportion of simple reflection and in-depth reflection.
Reflection on simple tasks emphasizes exploration based on valuable experience, while reflection on difficult tasks emphasizes the infusion of external failure lessons. The specific difficulty level assessment process can be expressed as:
\begin{equation}
DL=M_{complex}(type,num_{obj},num_{inter}),
\label{Flexible} 
\end{equation}
where $type$ represents the type of current task, $num_{obj}$ and $num_{inter}$ respectively represents the number of target objects and the number of interactions that need to be performed within the task. 
Among all episodes $ep_{total}$, the number of episodes for in-depth reflection is represented as:

\begin{equation}
k_{2}=ep_{total} \cdot \frac{num_{\text{obj}}+num_{\text{inter}}}{\max\left\{ num_{\text{obj}}(t) + num_{\text{inter}}(t) \,\bigg|\, t \in \mathcal{T} \right\}}
\end{equation}
and the number of episodes for simple reflection $k_{1}=ep_{total}-k_{2}$. 
The task difficulty assessment and detailed constructivism reflection process are illustrated in Figure~\ref{FCRF_detail}.

\section{EXPERIMENTS}

In this section, we evaluate our framework by conducting experiments in common household environment AlfWorld~\cite{shridhar2020alfworld}. Furthermore, we perform real-world robotic experiments to verify the practicability of our method in the real world environment. Our approach demonstrates significant advantages in overall performance and specific metrics. 

\subsection{Experimental Setup}



\textbf{Environment.} We conduct our evaluation on \textbf{AlfWorld}, a text-based virtual household environment containing six types (these types would be shown in next part). We test our framework and baselines on entire 134 tasks across all six types in five epochs of planning trails, demonstrating the superiority of our framework in terms of reflection performance and success rate through all methods. 

\textbf{Dataset.} In AlfWorld, we conduct experiments on the entire dataset including six types of household tasks: Pick \& Place, Examine in Light, Clean \& Place, Heat \& Place, Cool \& Place, and Pick Two \& Place. We perform experiments on all 134 tasks in the environment to obtain results. 

\textbf{Compared methods.} We compare our method with three categories of previous methods as main baselines on long-horizon planning tasks: 
1. \textbf{Planning-Only}: ProgPrompt~\cite{singh2023progprompt}, which LLM receives task descriptions in text form as input, and directly outputs action sequence as planning result based on the In Context Learning (ICL)~\cite{dong2022survey} principle. 
2. \textbf{Reasoning-Only}: ReAct~\cite{yao2022react}, which combines reasoning and acting processes with LLMs, generating the next action based on reasoning. 
3. \textbf{Reasoning-Reflection}: Reflexion~\cite{shinn2024reflexion}, which allows LLMs to reflect on their previous failures in ReAct trails according to environmental feedback, in order to form an improved plan for the next attempt. The SOTA method Expel~\cite{zhao2024expel} is also included in this category of self-reflection methods, while the overall success rate of our method exceeds it on up-to-date GPT-4 series models.

\textbf{Metrics.} We evaluated the methods from three aspects: 
\textbf{Success Rate} wholely evaluates the effectiveness of reflection and planning processes through calculating the proportion of tasks that are completed successfully to the end of trail.
\textbf{Flexibility} metric includes the Average Value (AVE) and Standard Deviation (STD) of generated self-reflection length, modeling the ability to manage reflection flexibly based on the difficulty of various tasks.
\textbf{Efficiency} metric includes redefined Experience Recall and Correction Precision during the self-reflection process in terms of action sequence level. 
The \textbf{Experience Recall} metrics the ability to retain valuable experience during the self-reflection process, which can be represented as $Recall_{exp} = \frac{C_{\text{retained}}}{C_{\text{initial}}}$, where $C_{retained}$ denotes the number of correct actions finally retained in the new plan that given by the reflection results, and $C_{initial}$ denotes the number of correct actions initially included in the ultimately failed action sequence. 
The \textbf{Correction Precision} measures the ability to accurately correct wrong steps during the self-reflection process, which can be represented as $Precision_{corr} = \frac{E_{\text{corrected}}}{E_{\text{total}}} $, where $E_{corrected}$ denotes the number of erroneous actions being corrected through the reflection process, and $E_{total}$ denotes the total number of erroneous actions in the ultimately failed action sequence.

\subsection{Main Results}

\begin{table*}[t]

    \centering
    \setlength{\tabcolsep}{0.9mm}
    \caption{Success rate of FCRF and baselines across various AlfWorld tasks. Our FCRF outperforms other methods on all tasks. }
    
\tabcolsep=0.02\linewidth
\begin{tabular}{lcccccc|c}
    \toprule
    \textbf{Methods} & \textbf{Put} & \textbf{Clean} & \textbf{Heat} & \textbf{Cool} & \textbf{Examine} & \textbf{Put two} & \textbf{ALL SR(\%)} \\
    
    \midrule
    \textbf{Planning-Only} & 64.5 & 66.6 & 30.0 & 62.5 & 26.6 & 35.7 & 68.6 \\
    \textbf{Reasoning-Only} & 84.6 & 87.5 & 77.1 & 93.3 & 58.3 & 88.8 & 82.8 \\
    \textbf{Reasoning-Reflection} & 76.9 & 84.3 & 80.0 & 93.3 & 58.3 & 96.2 & 83.5 \\
    \textbf{Ours} & \textbf{87.5} & \textbf{90.0} & \textbf{93.5} & \textbf{100} & \textbf{63.6} & \textbf{97.0} & \textbf{91.0} \\
   \midrule
\end{tabular}
\label{Table:main1}
\vspace{-2mm}
    
\end{table*}

\begin{table*}[t]

    \centering
    \setlength{\tabcolsep}{1.5mm}
    \caption{Flexibility and efficiency of different reflection methods. Our FCRF demonstrates significantly higher flexibility and efficiency under the condition of a 9.7\% increase in computational power consumption. }
\tabcolsep=0.015\linewidth
\begin{tabular}{lcccc}

    \toprule
    \multirow{2}{*}{\raisebox{-1ex}{\textbf{Methods}}} &
    \multicolumn{2}{c}{\textbf{Flexibility}} & \multicolumn{2}{c}{\textbf{Efficiency}} \\
    
    \cmidrule(lr){2-5}
    & AVE(words) & STD(words) & \textnormal{Recall$_{exp}$}(\%) & \textnormal{Precision$_{corr}$}(\%) \\
    \cmidrule(lr){1-5}
    
    \textbf{Reasoning-Reflection} & 371.1 & 199.9 & 75.0 & 32.1 \\
    \textbf{Ours} & \textbf{407.2} & \textbf{262.2} & \textbf{100.0} & \textbf{95.4} \\
    \midrule

\end{tabular}
    \label{Table:main2}
\vspace{-1mm}
\end{table*}

\begin{figure*}[!htbp]
\vspace{-2mm}
    \centering
    \includegraphics[width=0.99\textwidth]{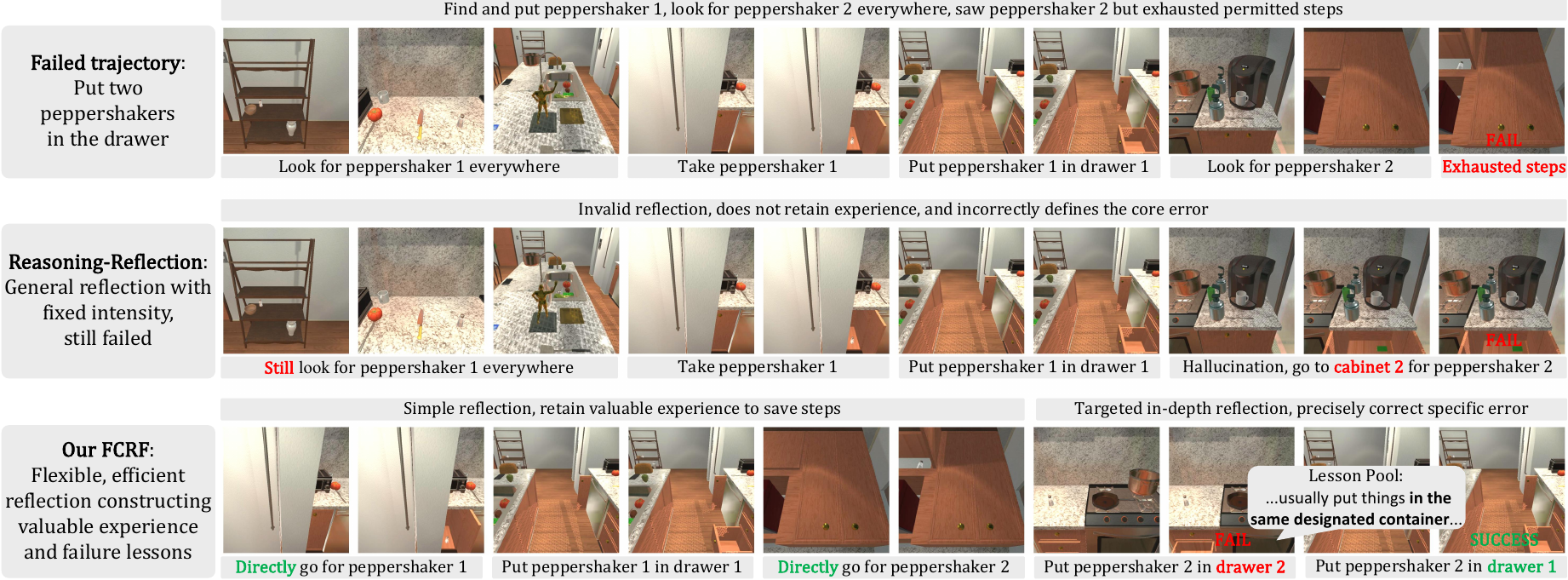}
    \caption{Comparison of reflection methods in an AlfWorld example. Faced with the failed trajectory, the Reasoning-Reflection method performs invalid reflection with inappropriate fixed intensity, which discards experience and fabricates a failure reason. While our FCRF first performs simple reflection, summarizing valuable experience to save steps, then FCRF precisely extractes lesson for the error under an in-depth reflection, finally corrects the trajectory. }
    \label{fig_sim_example}
    \vspace{-3mm}
\end{figure*}

The main results are presented in Table~\ref{Table:main1} and Table~\ref{Table:main2}. Balancing performance and computational resource consumption, all experiments are performed using the newest officially recommended \textit{GPT-4o mini} model. We can observe that: 
1. Our FCRF outperforms other methods on all tasks and metrics, denoting the ability of our method to enhance the overall success rate of planning, through performing flexible self-reflection according to task difficulty, and constructively integrates experience of both success and failure during the reflection process. An illustrative case of FCRF applied to a long-horizon task is shown in Figure~\ref{fig_sim_example}, which intuitively demonstrates the superiority of FCRF compared to the baseline method.
2. There is a noticeable trend that weaker self-reflection flexibility and efficiency correspond to an overall lower success rate, which demonstrates the necessity of our research topic and the rationality of metrics we defined.
3. Among all the evaluated methods, the \textbf{Planning-Only} method performs simple planning based on inputs and contexts, resulting in a relatively weakest overall success rate. 
The \textbf{Reasoning-Only} and \textbf{Reasoning-Reflection} methods achieve better results, with the overall performance of the Reasoning-Reflection method being slightly better than Reasoning-Only. However, in some task categories, Reasoning-Reflection performs worse than Reasoning-Only, demonstrating that self-reflection with fixed templates and strength does not always effectively correct planning errors.
4. Differently from the methods mentioned above, \textbf{our FCRF} can flexibly select the intensity of reflection based on the difficulty of the task and constructively integrate the experience of success and failure, thus broadening the overall success rate of varying difficulty tasks, compared to methods of all other categories, as shown in \textbf{Table~\ref{Table:main1}}.
More specifically, as shown in \textbf{Table~\ref{Table:main2}}, our volume of reflection contents varies more due to the difficulty of task, meanwhile, the average reflection length is only slightly higher than Reasoning-Reflection methods category, demonstrating the \textbf{flexibility} and superior cost-performance ratio of our framework. 
The \textbf{efficiency} metric shows that our approach has a higher effective experience recall and core error correction precision, demonstrating the capability of our method in constructing valuable experience and failure lessons, which benefits the overall efficacy of self-reflection.

\section{ANALYSIS AND DISCUSSION}

\subsection{Episode Analysis of Self-Reflection Process}


\begin{figure}
      \centering
      \includegraphics[width=0.5\textwidth]{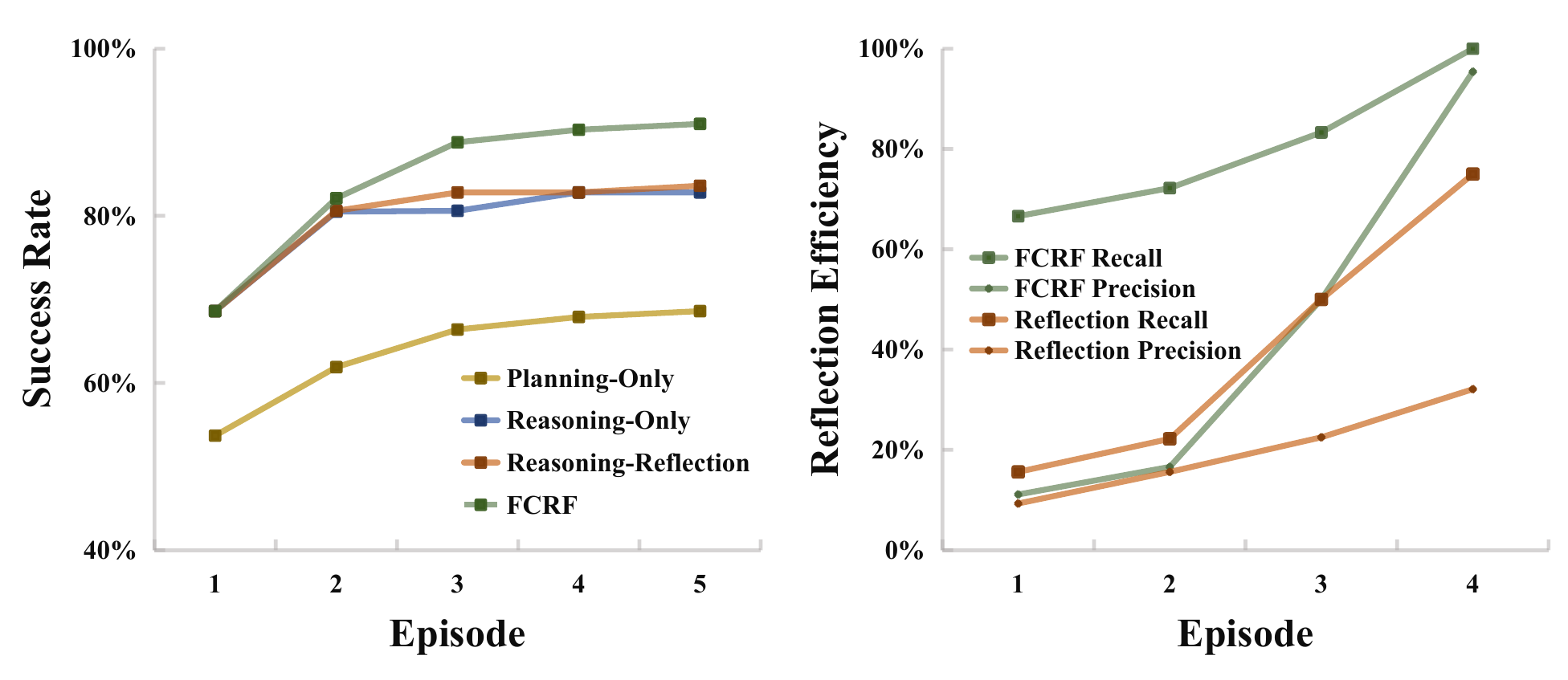}
      \caption{Result of success rate and reflection efficiency with different episodes. Our FCRF outperforms in all episodes, converges more quickly and has a significant better efficiency. }
    \label{EpisodeAnalysis}
    \vspace{-5mm}
\end{figure} 

We perform episode analysis experiments to specifically analyze the detailed mechanisms and manifestations of the self-reflection process, as shown in Figure~\ref{EpisodeAnalysis}. We instruct the LLM to continuously replan all erroneous long-horizon tasks over five rounds of experiments, and observe the completion status on these five episodes. Our observations are as follows.
1. As shown in the left image, our FCRF consistently outperforms the baselines in all episodes and converges more quickly, demonstrating that our method possesses superior performance and robustness together with a better computational cost efficiency. 
2. In all episodes, the simple Planning-Only method exhibits the weakest error correction performance, the Reasoning-Only and Reasoning-Reflection methods achieve performance improvements, but there is no noticeable gap between the two methods. 
3. As shown in the right image, our FCRF has a significantly stronger ability to recall valuable experience in the trajectory. After the lesson pool has been constructed online, our FCRF demonstrates significantly higher precision in error correction.
This further illustrates that reflection process with fixed intensity is not always necessarily conducive to error correction, while our implementation of reflection flexibility and the integration of valuable experience together with failure lessons are indeed conducive to the self-reflection process.





\subsection{Ablation Study}

\begin{table}[H]  
\vspace{-2mm}
    \centering
    \setlength{\tabcolsep}{1mm}
    \caption{Ablation of modules of FCRF in AlfWorld tasks. All designed modules contribute to the performance.}
\begin{tabular}{lccccc}

\toprule
& SR & AVE & STD & \textnormal{Recall$_{exp}$}(\%) & \textnormal{Precision$_{corr}$}(\%) \\
\midrule
\textbf{Ours Full} & 91.0 & 407.2 & 262.2 & 100.0 & 95.4 \\
\textbf{w/o Experience} & 87.3 & 238.4 & 82.8 & 37.5 & 88.9 \\
\textbf{w/o Lesson} & 88.8 & 276.2 & 110.7 & 100.0 & 20.0 \\
\bottomrule
\end{tabular}
    \label{Table:ablation}
    \vspace{-3.5mm}
\end{table}

To fully demonstrate the effectiveness of each module in our framework and to explore the interrelationships between them, we perform ablation studies on the full AlfWorld dataset, and the results are detailed in Table~\ref{Table:ablation}. 
In the \textbf{w/o Experience} model, the action sequence is planned without extracting valuable experience. Compared to the complete model, the overall success rate decreases by 3.7\%. The ability to recall valuable experience obviously reduces to 37.5\%, while the ability to precisely correct errors still exists. 
In the \textbf{w/o Lesson} model, the action sequence is planned without using failure lesson, the success rate decreases by 2.2\%. The ability to precisely correct errors reduces to 20\% while the ability to recall the experience is still great. 
The flexibility of the method decreases when each module is ablated, leading to a decrease of overall success rate, but the performance of each ablated model is still better than other baselines. 
Altogether, results of the ablation study met expectations on each metric, demonstrating the effect of each module in our method, and showing that our framework the whole is greater than the simple sum of its parts.

\subsection{The Extracted Lesson Pool Result}
\begin{figure}[H]
\vspace{-1mm}
      \centering
      \includegraphics[width=0.47\textwidth]{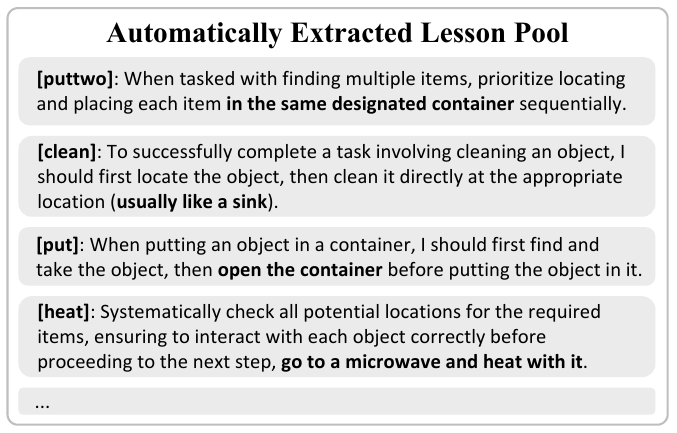}
      \caption{Part of the automatically extracted lesson pool by the Mentor LLM in our experiment. The lesson pool may contain environmental constraints with the significance of retention and generalization. }
    \label{fig_lesson_pool}
    \vspace{-2mm}
\end{figure}

In our Mentor-Actor reflection architecture, as mentioned in methodology section, the Mentor LLM visits successfully corrected trajectories of the Actor LLM through an online process, extracting a lesson pool to guide the subsequent reflection. 
The automatically extracted lesson pool is equivalent to a generalized knowledge base in task scenarios, which may contain environmental constraints that humanity has not yet recognized, thus possesses the significance of retention and generalization. The lesson pool obtained during our experimental process is partly displayed in Figure~\ref{fig_lesson_pool}.

\subsection{Real-World Robotic Experiment}
\begin{figure*}[!htbp]
      \centering
      \includegraphics[width=0.8\textwidth]{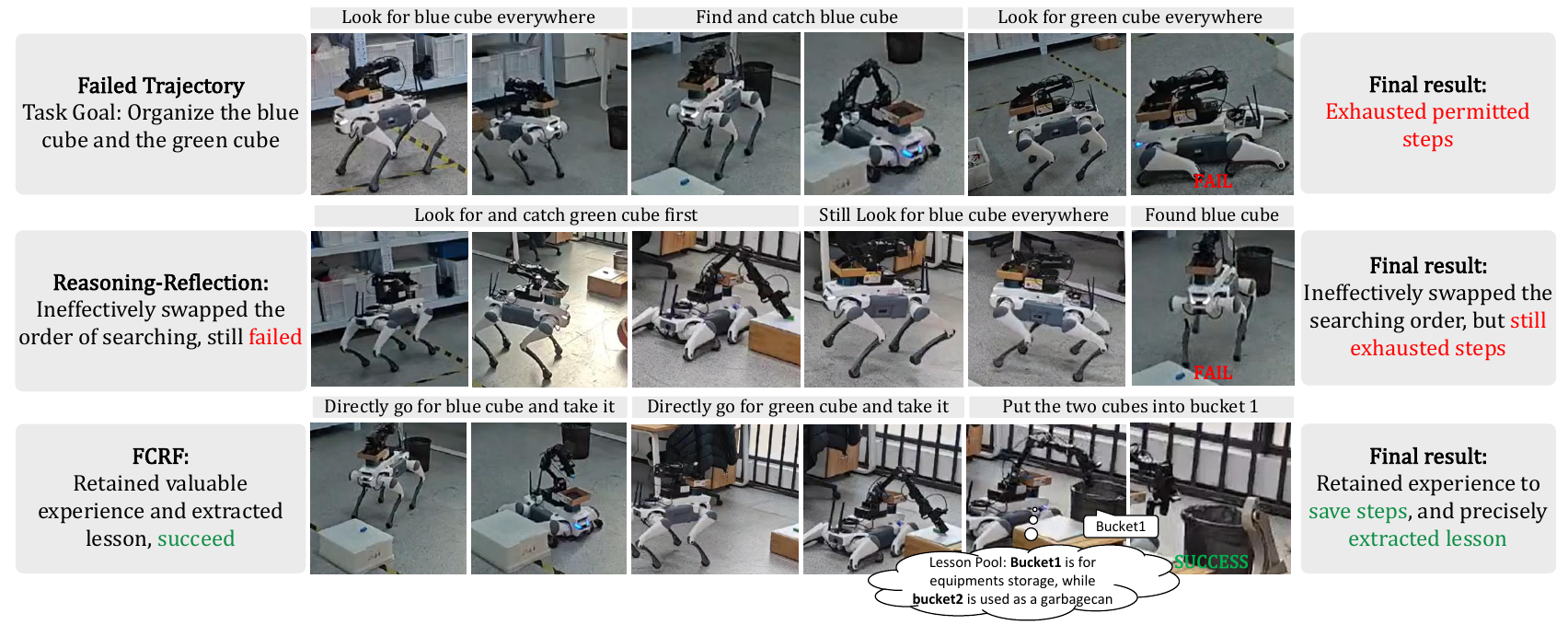}
      \caption{Comparison of reflection methods in a real-world experiment. In the block organization task, Reasoning-Reflection invalidly swaps the order of searching, while FCRF retains experience of object position to save action steps and extracts precise lesson for constraints, finally corrected the trajectory. }
    \label{real_world}
    \vspace{-3mm}
\end{figure*}

We use a quadruped robot with a manipulator to validate the practicability of our method in the real world. The results indicate that the robot deployed with our FCRF performs better in self-reflection on a block organization task. The detailed process is shown in Figure~\ref{real_world}. 
More details of real-world experiments will be shown in our website and video.

\section{CONCLUSIONS}
To our knowledge, we are the first to study the problem of self-reflection flexibility and constructivism in long-horizon robotic task planning with LLMs. Based on the constructivist learning theory of human intelligence, we propose a Mentor-Actor self-reflection framework called FCRF, which performs self-reflection with intensity flexibility according to task difficulty, meanwhile integrating valuable experience and failure lessons. Furthermore, we design a memory management module to efficiently manage the reflection context of LLMs. Experiments conducted in virtual household environment and the real world demonstrate that our reflection framework can effectively enhance the flexibility, efficiency and success rate of long-horizon planning tasks, making domestic robots more reliable and trustworthy.






\bibliographystyle{unsrt}
\bibliography{main}

\begin{thebibliography}{10}

\bibitem{guo2023recent}
Huihui Guo, Fan Wu, Yunchuan Qin, Ruihui Li, Keqin Li, and Kenli Li.
\newblock Recent trends in task and motion planning for robotics: A survey.
\newblock {\em ACM Computing Surveys}, 55(13s):1--36, 2023.

\bibitem{zachiotis2018survey}
Georgios~A Zachiotis, George Andrikopoulos, Randy Gornez, Keisuke Nakamura, and George Nikolakopoulos.
\newblock A survey on the application trends of home service robotics.
\newblock In {\em 2018 IEEE international conference on Robotics and Biomimetics (ROBIO)}, pages 1999--2006. IEEE, 2018.

\bibitem{ahn2022can}
Michael Ahn, Anthony Brohan, Noah Brown, Yevgen Chebotar, Omar Cortes, Byron David, Chelsea Finn, Chuyuan Fu, Keerthana Gopalakrishnan, Karol Hausman, et~al.
\newblock Do as i can, not as i say: Grounding language in robotic affordances.
\newblock {\em arXiv preprint arXiv:2204.01691}, 2022.

\bibitem{kannan2024smart}
Shyam~Sundar Kannan, Vishnunandan~LN Venkatesh, and Byung-Cheol Min.
\newblock Smart-llm: Smart multi-agent robot task planning using large language models.
\newblock In {\em 2024 IEEE/RSJ International Conference on Intelligent Robots and Systems (IROS)}, pages 12140--12147. IEEE, 2024.

\bibitem{zhou2024isr}
Zhehua Zhou, Jiayang Song, Kunpeng Yao, Zhan Shu, and Lei Ma.
\newblock Isr-llm: Iterative self-refined large language model for long-horizon sequential task planning.
\newblock In {\em 2024 IEEE International Conference on Robotics and Automation (ICRA)}, pages 2081--2088. IEEE, 2024.

\bibitem{yao2023retroformer}
Weiran Yao, Shelby Heinecke, Juan~Carlos Niebles, Zhiwei Liu, Yihao Feng, Le~Xue, Rithesh Murthy, Zeyuan Chen, Jianguo Zhang, Devansh Arpit, et~al.
\newblock Retroformer: Retrospective large language agents with policy gradient optimization.
\newblock {\em arXiv preprint arXiv:2308.02151}, 2023.

\bibitem{shinn2024reflexion}
Noah Shinn, Federico Cassano, Ashwin Gopinath, Karthik Narasimhan, and Shunyu Yao.
\newblock Reflexion: Language agents with verbal reinforcement learning.
\newblock {\em Advances in Neural Information Processing Systems}, 36, 2024.

\bibitem{yao2022react}
Shunyu Yao, Jeffrey Zhao, Dian Yu, Nan Du, Izhak Shafran, Karthik Narasimhan, and Yuan Cao.
\newblock React: Synergizing reasoning and acting in language models.
\newblock {\em arXiv preprint arXiv:2210.03629}, 2022.

\bibitem{yao2024tree}
Shunyu Yao, Dian Yu, Jeffrey Zhao, Izhak Shafran, Tom Griffiths, Yuan Cao, and Karthik Narasimhan.
\newblock Tree of thoughts: Deliberate problem solving with large language models.
\newblock {\em Advances in Neural Information Processing Systems}, 36, 2024.

\bibitem{zhang2024agent}
Wenqi Zhang, Ke~Tang, Hai Wu, Mengna Wang, Yongliang Shen, Guiyang Hou, Zeqi Tan, Peng Li, Yueting Zhuang, and Weiming Lu.
\newblock Agent-pro: Learning to evolve via policy-level reflection and optimization.
\newblock {\em arXiv preprint arXiv:2402.17574}, 2024.

\bibitem{zhou2023language}
Andy Zhou, Kai Yan, Michal Shlapentokh-Rothman, Haohan Wang, and Yu-Xiong Wang.
\newblock Language agent tree search unifies reasoning acting and planning in language models.
\newblock {\em arXiv preprint arXiv:2310.04406}, 2023.

\bibitem{zhao2024expel}
Andrew Zhao, Daniel Huang, Quentin Xu, Matthieu Lin, Yong-Jin Liu, and Gao Huang.
\newblock Expel: Llm agents are experiential learners.
\newblock In {\em Proceedings of the AAAI Conference on Artificial Intelligence}, volume~38, pages 19632--19642, 2024.

\bibitem{chen2024automanual}
Minghao Chen, Yihang Li, Yanting Yang, Shiyu Yu, Binbin Lin, and Xiaofei He.
\newblock Automanual: Generating instruction manuals by llm agents via interactive environmental learning.
\newblock {\em arXiv preprint arXiv:2405.16247}, 2024.

\bibitem{hein1991constructivist}
George~E Hein.
\newblock Constructivist learning theory.
\newblock {\em Institute for Inquiry. Available at:/http://www. exploratorium. edu/ifi/resources/constructivistlearning. htmlS}, 1991.

\bibitem{shridhar2020alfworld}
Mohit Shridhar, Xingdi Yuan, Marc-Alexandre C{\^o}t{\'e}, Yonatan Bisk, Adam Trischler, and Matthew Hausknecht.
\newblock Alfworld: Aligning text and embodied environments for interactive learning.
\newblock {\em arXiv preprint arXiv:2010.03768}, 2020.

\bibitem{li2022pre}
Shuang Li, Xavier Puig, Chris Paxton, Yilun Du, Clinton Wang, Linxi Fan, Tao Chen, De-An Huang, Ekin Aky{\"u}rek, Anima Anandkumar, et~al.
\newblock Pre-trained language models for interactive decision-making.
\newblock {\em Advances in Neural Information Processing Systems}, 35:31199--31212, 2022.

\bibitem{radford2019language}
Alec Radford, Jeffrey Wu, Rewon Child, David Luan, Dario Amodei, Ilya Sutskever, et~al.
\newblock Language models are unsupervised multitask learners.
\newblock {\em OpenAI blog}, 1(8):9, 2019.

\bibitem{liang2023code}
Jacky Liang, Wenlong Huang, Fei Xia, Peng Xu, Karol Hausman, Brian Ichter, Pete Florence, and Andy Zeng.
\newblock Code as policies: Language model programs for embodied control.
\newblock In {\em 2023 IEEE International Conference on Robotics and Automation (ICRA)}, pages 9493--9500. IEEE, 2023.

\bibitem{singh2023progprompt}
Ishika Singh, Valts Blukis, Arsalan Mousavian, Ankit Goyal, Danfei Xu, Jonathan Tremblay, Dieter Fox, Jesse Thomason, and Animesh Garg.
\newblock Progprompt: Generating situated robot task plans using large language models.
\newblock In {\em 2023 IEEE International Conference on Robotics and Automation (ICRA)}, pages 11523--11530. IEEE, 2023.

\bibitem{sun2024adaplanner}
Haotian Sun, Yuchen Zhuang, Lingkai Kong, Bo~Dai, and Chao Zhang.
\newblock Adaplanner: Adaptive planning from feedback with language models.
\newblock {\em Advances in Neural Information Processing Systems}, 36, 2024.

\bibitem{song2023llm}
Chan~Hee Song, Jiaman Wu, Clayton Washington, Brian~M Sadler, Wei-Lun Chao, and Yu~Su.
\newblock Llm-planner: Few-shot grounded planning for embodied agents with large language models.
\newblock In {\em Proceedings of the IEEE/CVF International Conference on Computer Vision}, pages 2998--3009, 2023.

\bibitem{wake2023chatgpt}
Naoki Wake, Atsushi Kanehira, Kazuhiro Sasabuchi, Jun Takamatsu, and Katsushi Ikeuchi.
\newblock Chatgpt empowered long-step robot control in various environments: A case application.
\newblock {\em IEEE Access}, 2023.

\bibitem{lauri2022partially}
Mikko Lauri, David Hsu, and Joni Pajarinen.
\newblock Partially observable markov decision processes in robotics: A survey.
\newblock {\em IEEE Transactions on Robotics}, 39(1):21--40, 2022.

\bibitem{zhang2024fltrnn}
Jiatao Zhang, Lanling Tang, Yufan Song, Qiwei Meng, Haofu Qian, Jun Shao, Wei Song, Shiqiang Zhu, and Jason Gu.
\newblock Fltrnn: Faithful long-horizon task planning for robotics with large language models.
\newblock In {\em 2024 IEEE International Conference on Robotics and Automation (ICRA)}, pages 6680--6686. IEEE, 2024.

\bibitem{dong2022survey}
Qingxiu Dong, Lei Li, Damai Dai, Ce~Zheng, Jingyuan Ma, Rui Li, Heming Xia, Jingjing Xu, Zhiyong Wu, Tianyu Liu, et~al.
\newblock A survey on in-context learning.
\newblock {\em arXiv preprint arXiv:2301.00234}, 2022.

\end{thebibliography}

\end{document}